\documentclass[]{article}

\usepackage{cite}
\usepackage{amsmath,amssymb,amsfonts}
\usepackage{algorithmic}
\usepackage{graphicx}
\usepackage{textcomp}
\usepackage[table]{xcolor}
\usepackage{marvosym}
\usepackage{makecell}
\usepackage{multirow}
\usepackage[ruled,vlined,linesnumbered]{algorithm2e}
\usepackage[shortcuts]{extdash}
\usepackage[normalem]{ulem}
\usepackage{subcaption}
\usepackage{todonotes}
\usepackage{tabu}
\usepackage{booktabs}

\usepackage[margin=40mm]{geometry}

\providecommand{\keywords}[1]
{
  \small	
  \textbf{\textit{Keywords---}} #1
}

\begin{document}

\title{Prediction of the facial growth direction with\\ Machine Learning methods}
%
%\titlerunning{Abbreviated paper title}
% If the paper title is too long for the running head, you can set
% an abbreviated paper title here
%
\author{
Stanisław Kaźmierczak$^1$,
Zofia Juszka$^2$,\\
Piotr Fudalej$^{3,4,5}$,
Jacek Ma{\'n}dziuk$^1$ \\ \\

\small $^1$Faculty of Mathematics and Information Science, Warsaw University of Technology, \\
\small Warsaw, Poland\\
\small \texttt{\{s.kazmierczak,mandziuk\}@mini.pw.edu.pl} \\
\small $^2$Prof. Loster's Orthodontics, ul. Bartłomieja Nowodworskiego 4, \\
\small 30-433 Krakow, Poland \\ 
\small \texttt{zofia.juszka@gmail.com} \\
\small $^3$Department of Orthodontics, Jagiellonian University in Krakow, Krakow, Poland \\
\small \texttt{pfudalej@gmail.com} \\
\small $^4$Department of Orthodontics, Institute of Dentistry and Oral Sciences, \\
\small Palacký University Olomouc, Olomouc, Czech Republic \\
\small $^5$ Department of Orthodontics and Dentofacial Orthopedics, \\ \small University of Bern, Bern, Switzerland
}

% \author{Author Surname$^{1}$, Someone Else$^{2}$  \\
%         \small $^{1}$Faculty of Mathematics and Information Science, Warsaw University of Technology, Warsaw, Poland \\
%         \small $^{2}$University B \\
% }
\date{} % Comment this line to show today's date

% \institute{Faculty of Mathematics and Information Science, Warsaw University of Technology, Warsaw, Poland \\
% \email{\{s.kazmierczak,mandziuk\}@mini.pw.edu.pl}
% \and
% Prof. Loster's Orthodontics, ul. Bartłomieja Nowodworskiego 4, 30-433 Krakow, Poland \\ \email{zofia.juszka@gmail.com} 
% \and
% Department of Orthodontics, Jagiellonian University in Krakow, Krakow, Poland
% \email{pfudalej@gmail.com} \and
%  Department of Orthodontics, Institute of Dentistry and Oral Sciences, Palacký University Olomouc, Olomouc, Czech Republic \and Department of Orthodontics and Dentofacial Orthopedics, University of Bern, Bern, Switzerland}

\maketitle              % typeset the header of the contribution

\begin{abstract}
First attempts of prediction of the facial growth (FG) direction were made over half of a century ago. Despite numerous attempts and elapsed time, a satisfactory method has not been established yet and the problem still poses a challenge for medical experts. To our knowledge, this paper is the first Machine Learning approach to the prediction of FG direction. Conducted data analysis reveals the inherent complexity of the problem and explains the reasons of difficulty in FG direction prediction based on 2D X-ray images. To perform growth forecasting, we employ a wide range of algorithms, from logistic regression, through tree ensembles to neural networks and consider three, slightly different, problem formulations. The resulting classification accuracy varies between 71\% and 75\%.

\end{abstract}

%%%%%%%%%%%%%%%%%%%% Introduction %%%%%%%%%%%%%%%%%%%%
\section{Introduction}
\label{sec:Introduction}
Characteristics of FG -- direction, intensity, and duration - can be critical in therapy of congenital malformations and malocclusions. From the clinical perspective, FG can be viewed as favorable, neutral, or unfavorable. Favorable growth occurs when components of the face grow in a direction or with intensity facilitating treatment, while unfavorable growth takes place when growth characteristics hinders achievement of optimal results of treatment. Neutral growth is located between these extremes. 

Over the last several decades, FG was studied primarily with two-dimensional (2D) radiography. To this end "growth studies" i.e., collections of girls and boys in whom profile 2D radiographs of the head had been taken annually, were initiated in the USA in the 1930's. Until the late 1980's when it became gradually clear that exposure of humans to ionizing radiation only for scientific purposes was unethical, thus unacceptable, more than 2000 children and teenagers had been followed.

Characterization of FG on 2D radiographs still poses a significant challenge even to an experienced investigator/clinician. It is due to paucity of stable reference structures (all parts of the face grow and remodel) and temporally and spatially diversified growth of individual components of the face (i.e. different parts of the face can grow at maximum velocity at different ages and different parts of the face can grow at different directions at different ages, respectively). Furthermore, direction of overall FG can vary depending on the age or developmental period.

Another challenge in assessment of FG is the lack of coordinate system within which the direction of growth can be unambiguously determined. Both in research and clinical practice, different angular or linear measurements or proportions are used for facial classification. Unfortunately, the facial morphology or growth direction categorized as 'X' by a given measurement (labeling) can fall into a different category when alternative measurement (labeling) is applied.

The main contribution of this work is threefold:
\begin{itemize}
    \item addressing the problem of prediction of the growth of the face with an application of machine learning (ML) methods, which, to the best of our knowledge, has not been analyzed before;
    \item performing a comprehensive data analysis which revealed the complexity of the problem and showed why a prediction based on 2D X-ray images is truly challenging;
    \item applying and comparing a broad set of ML algorithms, best of which achieved 71\% to 75\% classification accuracy.
\end{itemize}
% The remainder of this paper is arranged as follows. Section~\ref{sec:RelatedLiterature} provides a literature review. Section~\ref{sec:Datasets} characterizes datasets. Sections~\ref{sec:DataAnalysis} and~\ref{sec:Experiments} describe the data analysis and conducted prediction experiments along with their results, respectively. Finally, brief conclusions and directions for further research are presented in the last section.

%%%%%%%%%%%%%%%%%%%% Related literature %%%%%%%%%%%%%%%%%%%%
\section{Related literature}
\label{sec:RelatedLiterature}

First attempts of prediction of FG were made in 1960's and 1970's -- the authors of~\cite{bjork1969prediction} proposed a structural method consisting in qualitative assessment of seven morphological features on a single 2D radiograph, while in~\cite{bhatia1979proposed} statistical modeling was used for classification of facial types, hence FG at 9 and 17 years of age. Later, the authors of~\cite{skieller1984prediction} proposed a multivariate regression model with four morphological features as having $86\%$ predictive estimate of mandibular growth rotation. In~\cite{buschang1990mandibular}, the authors developed a prediction system by adding mean annual velocities with predictions derived from a polynomial model of the population's growth curve, while the authors of~\cite{rudolph1998multivariate} created prediction equations for identification of favorable vs. unfavorable growth patterns. Unfortunately, a possibility of effective FG prediction was questioned by others (\hspace{1sp}\cite{baumrind1984prediction}, \cite{lux1999evaluation}, \cite{kolodziej2002evaluation}) who observed that when previously proposed methods were used in different patient samples, their predictive power was limited and clinically irrelevant.

An interesting alternative in the search for the method of prediction of FG can be application of machine learning (ML), which has not been used to this end yet. Until now, methods of ML have been recently applied for automated identification of cephalometric landmarks on 2D radiographs (\hspace{1sp}\cite{arik2017fully}, \cite{park2019automated}, \cite{hwang2020automated}).
% or on 3D computed tomographic (CT) images (\hspace{1sp}\cite{ma2020automatic})
In~\cite{arik2017fully}, the authors used deep convolutional neural networks (CNNs) for fully automated quantitative cephalometric analysis, the authors of~\cite{park2019automated}, compared two newly developed deep learning algorithms for automatic identification of cephalometric landmarks annotated on 2D radiographs -- You-Only-Look-Once version 3 (YOLOv3) and Single Shot Multibox Detector (SSD), while in~\cite{hwang2020automated} performance of human experts and YOLOv3 algorithm for automatic identification of cephalometric landmarks was compared. 
% The authors of~\cite{ma2020automatic}, in turn, developed an automatic landmarking model based on a patch-based deep neural network for multiple landmark localization in CT images. 

In summary, application of ML methods in clinical disciplines related to treatment of craniofacial anomalies is  growing fast in recent years. Currently, some methods of automatic identification of cephalometric landmarks can be as accurate as those of human experts. However, to our knowledge, no research on the possibility of FG prediction with ML methods has been presented yet.

%%%%%%%%%%%%%%%%%%%% Datasets %%%%%%%%%%%%%%%%%%%%
\section{Datasets}
\label{sec:Datasets}

Assessment of FG requires a collection of longitudinal (i.e., taken at fixed intervals) images of the same subject. The images should comprehensively capture 3D facial morphology (3D images are preferred to 2D images). Then, images should be made in a standardized fashion ensuring comparability 'along' individual subject and across a sample and they should be made at standardized ages or developmental period. Fulfilling these requirements is challenging not only because of practical issues such as frequent drop-outs of subjects from the studies  but also because of ethical issues (eg., taking 3D radiographic images at fixed intervals in healthy subjects is considered unethical and forbidden; as a result, only already existing 2D radiographs can be studied). Therefore, participants from Norwegian "Nittedal Growth Study"~\cite{el1994longitudinal} and AAOF Craniofacial Growth Legacy Collection (collections included: Bolton-Brush, Burlington, Denver, Fels Longitudinal, Forsyth Twin, Iowa, Mathews, Michigan, Oregon) ~\cite{american1996growth} were included in this investigation. In summary, children born between the 1950's and mid-1970's residing within a specific geographic area of the respective Growth Study reported for a clinical examination and 2D radiographic registration at predetermined ages (intervals). It should be emphasized that imaging technique and age at radiographic imaging were attempted to be standardized only within the individual Growth Study but not between the Studies. As a result, radiographs taken at different locations can differ in image magnification or the age of imaging.

When preparing the dataset, we aimed at having data from 9- and 12-year-olds as an input to predict the change of particular measurements between the 9th and 18th year of age. However, due to a lack of sufficient data, we decided to make these assumptions less strict (to be included in the group of 9-year-olds, one is not obliged to be exactly 9 years old, etc.) to gather enough instances. The total number of samples amounts to 639. Characteristics of particular groups present Table~\ref{tab:datasetCharacteristic}.
\begin{table}[t]
\centering
\caption{Characteristic of 9-, 12- and 18-year-olds.} \label{tab:datasetCharacteristic}
\begin{tabular}{m{2.2cm}m{2.2cm}m{1.5cm}m{1.5cm}}
\toprule
Group & Age & Minimum & Maximum \\
\midrule
\emph{9-year-olds} & $9.06 \pm 0.45$ & 6.00 & 10.92 \\
\emph{12-year-olds} & $12.07 \pm 0.39$ & 10.00 & 13.75\\
\emph{18-year-olds} & $17.41 \pm 1.71$ & 15.00 & 28.42\\
\bottomrule
\end{tabular}
\end{table}

%%%%%%%%%%%%%%%%%%%% Data analysis %%%%%%%%%%%%%%%%%%%%
\section{Data analysis}
\label{sec:DataAnalysis}

\subsection{Landmarking}
\label{sec:DataAnalysis:Landmarking}
Instances of FG direction are in the form of tabular data. The feature set is composed of cephalometric variables or different coordinates described later. Irrespective of the feature format, an orthodontic expert has to identify about twenty characteristic landmarks on each radiograph of the face. The location of some landmarks is ambiguous and it is impossible to mark such points very precisely \cite{perillo2000effect}. Thus, the data gathering process poses a source of noise.

\begin{figure}[t]
     \centering
     \begin{minipage}[t]{0.38\textwidth}
         \subfloat[] 
         {\includegraphics[width=\textwidth]{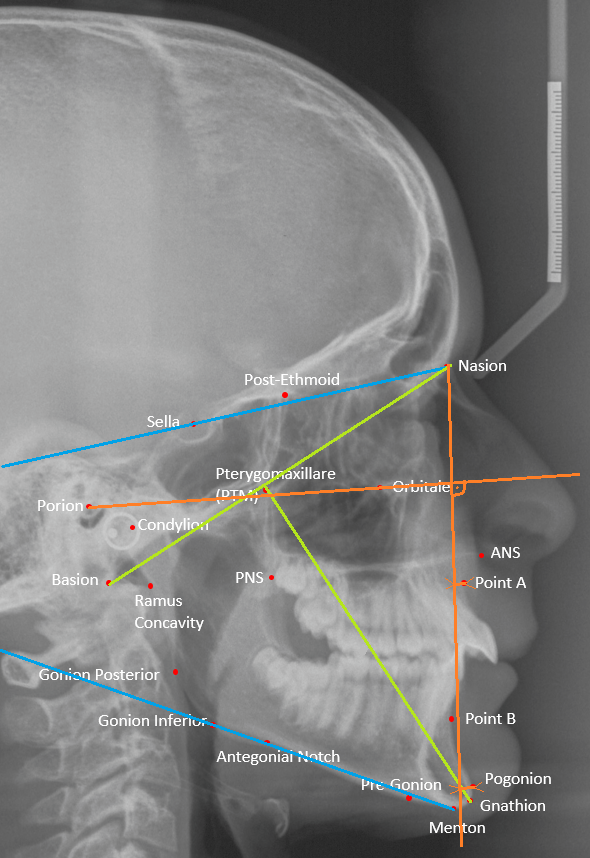}}
     \end{minipage}
     \hfill
     \begin{minipage}[b]{0.6\textwidth}
        \subfloat[]
        {\includegraphics[width=\textwidth]{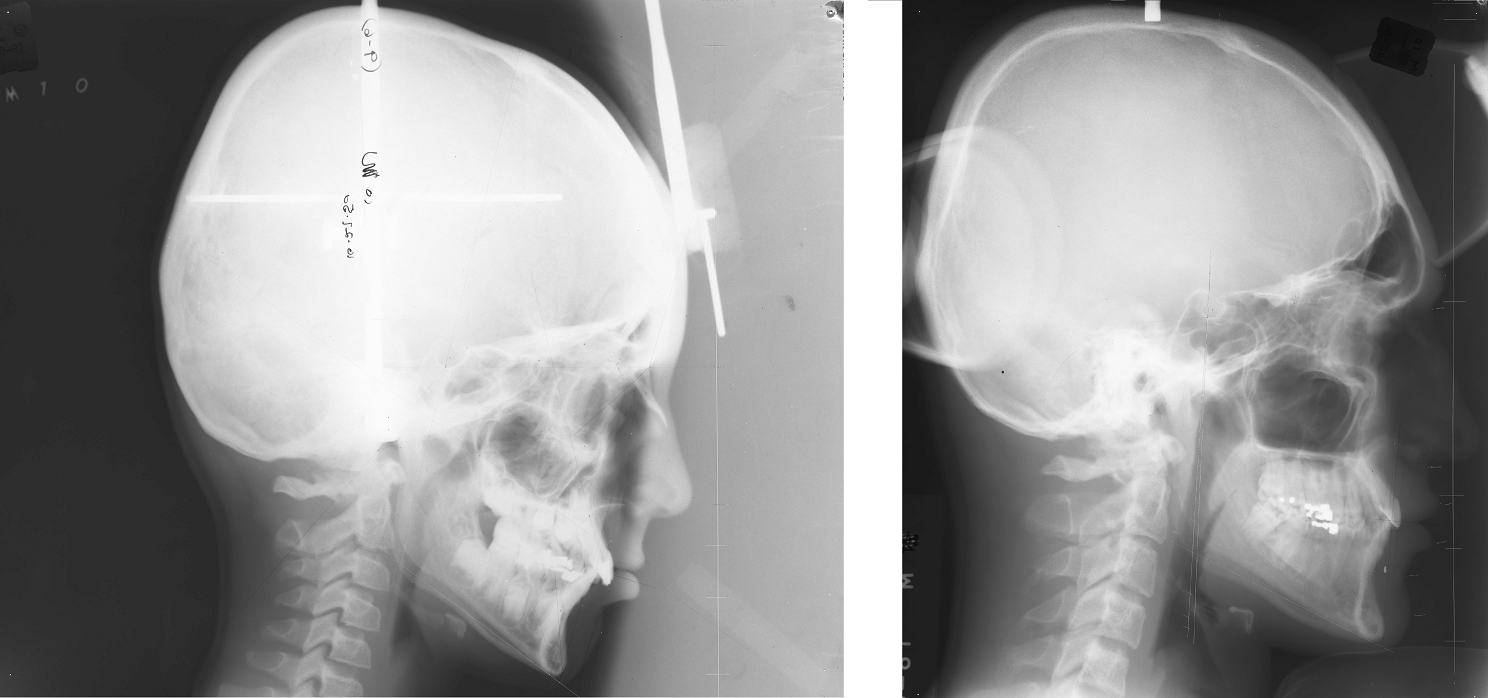}}\\
        \subfloat[]
        {\includegraphics[width=\textwidth]{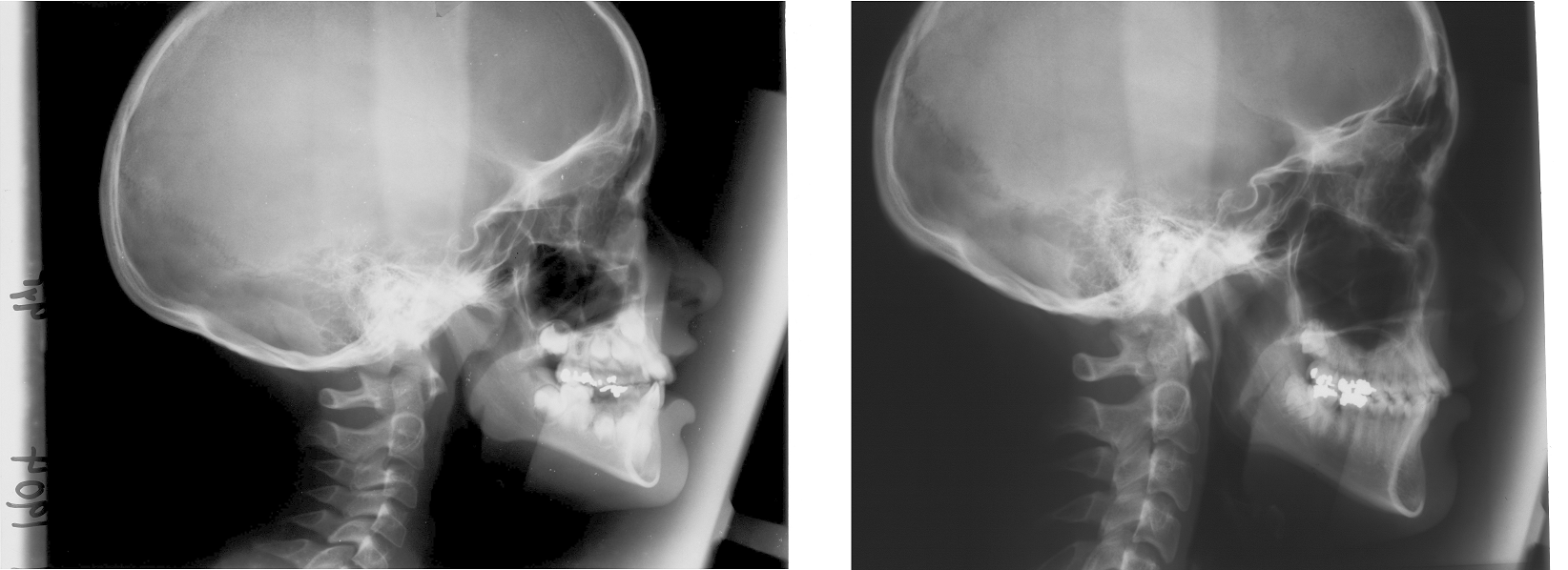}}
    \end{minipage}
    \caption{Sample cephalograms. Figure (a) depicts marked landmarks, their names, as well as measurements used to  create  the  predicted variables: SN/MP (blue), FA (green), PN-AN (orange). Figures (b) and (c) present horizontal and vertical growth, respectively. They illustrate the faces of two people in the age of 9 (left) and 18 (right). }
    \label{fig:RTGExamples}
\end{figure}

Although there is no widely accepted set of measurements to assess FG direction, orthodontists can utilize the following \emph{three central measurements}: SN/MP, FA, and PN-AN.  Each of them corresponds to a slightly different aspect of growth. SN/MP is an angle between Sella-Nasion and Menton-Gonion Inferior lines. FA (facial axis) is an angle formed by Basion-Nasion and Gnathion-PTM. PN-AN is defined by the difference of distances from Point A and Pogonion to the line perpendicular to Porion-Orbitale and going through Nasion. FG direction can be described by one of three categories: \emph{horizontal, vertical, and mixed}. Fig.~\ref{fig:RTGExamples}a depicts sample X-ray photographs with landmarks and measurements used to create the predicted variables, while Fig.~\ref{fig:RTGExamples}b and Fig.~\ref{fig:RTGExamples}c some examples of horizontal and vertical growth.

\subsection{Face size}
\label{sec:DataAnalysis:InitialTransformations}
\begin{figure}[b!]
\centerline{\includegraphics[width=0.64\textwidth]{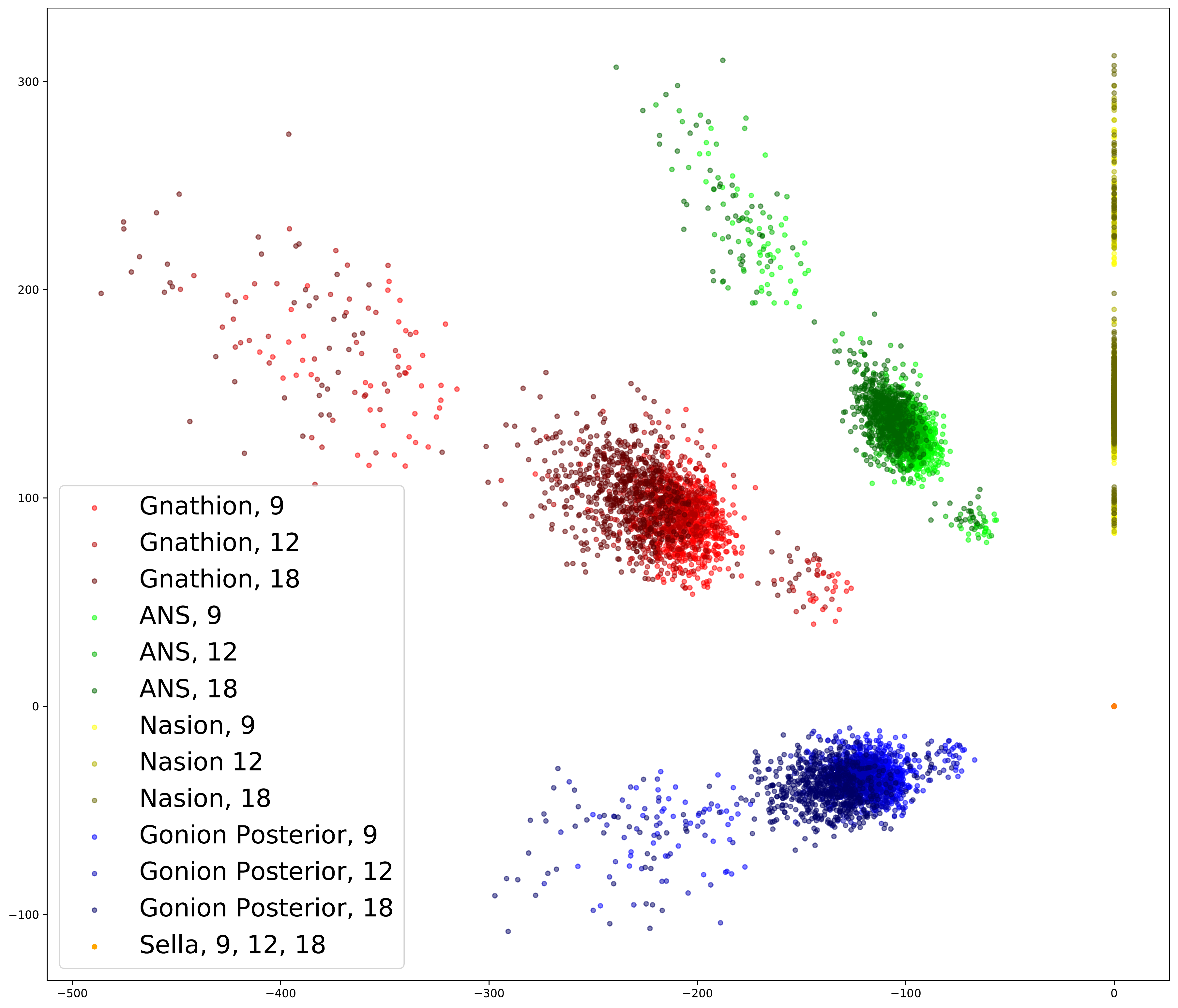}}
\caption{Cloud of selected five landmark coordinates coming from all patients at the ages of 9, 12 and 18.}
\label{fig:DifferentScales}
\end{figure}
The range of coordinates is extensive. Fig.~\ref{fig:DifferentScales} depicts clouds of landmark coordinates coming from all patients at the age of 9, 12, and 18. For the sake of readability, we limited the number of presented points to five and rotated them to make Sella-Nasion line vertical. The visualization shows that the images were taken at least in three or four different scales. It is not possible to retrieve any absolute measurement values from images. It means that one objectively smaller face can be larger in image than the other one and thus larger in the coordinates space. It may concern images of different people but, worse still, images of one person which confuse a model. Precise scaling is not possible since we do not know which value of the scaling factor to apply. 

\subsection{Comparison of patients groups}
\label{sec:DataAnalysis:GroupComparison}
\begin{figure}[!b]
    \captionsetup[subfigure]{justification=centering}
     \centering
     \begin{subfigure}[]{0.38\textwidth}
         \centering
         \includegraphics[width=\textwidth]{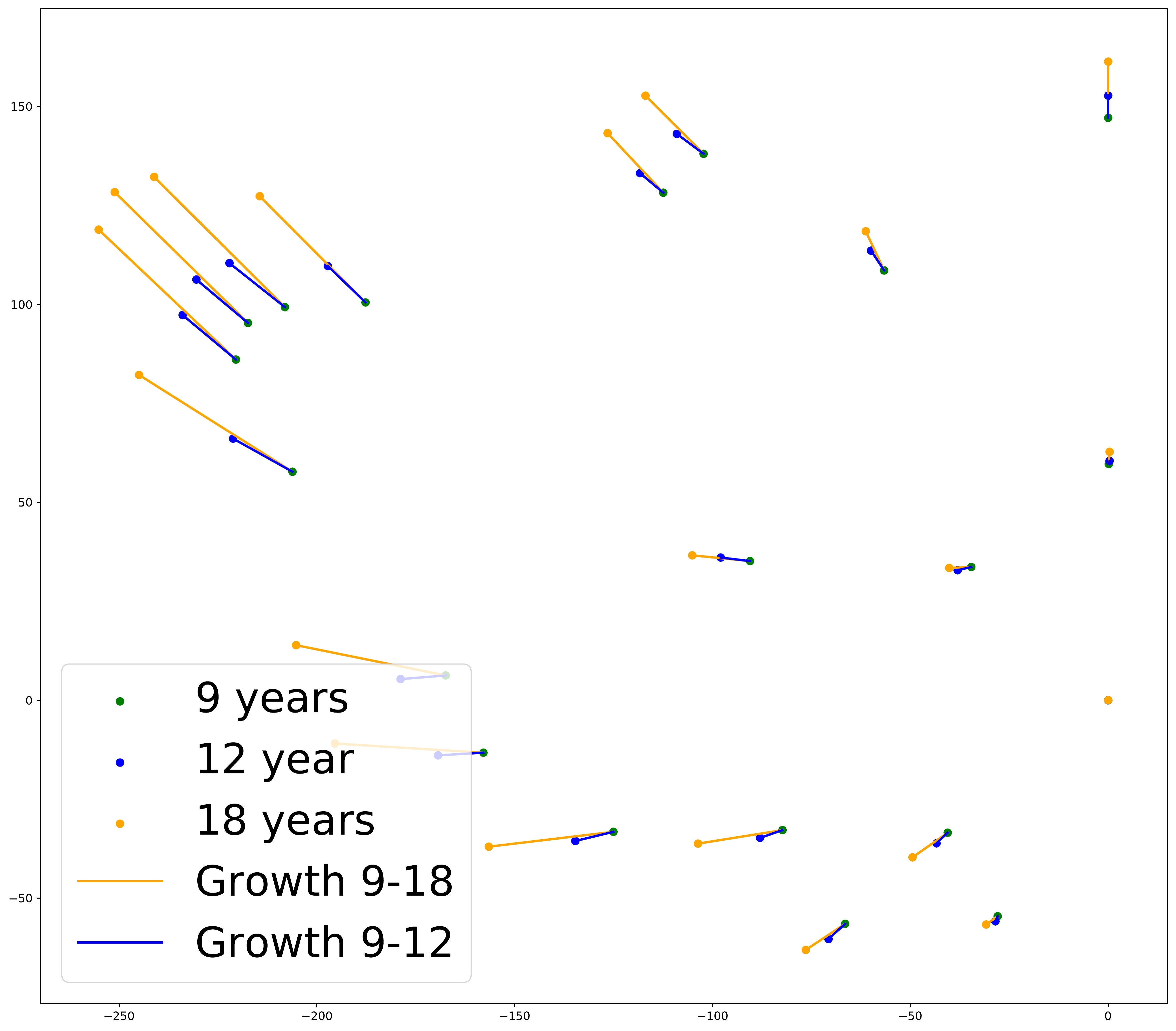}
         \caption{}
         \label{subfig:forwardGrowth}
     \end{subfigure}
     \begin{subfigure}[]{0.385\textwidth}
         \centering
         \includegraphics[width=\textwidth]{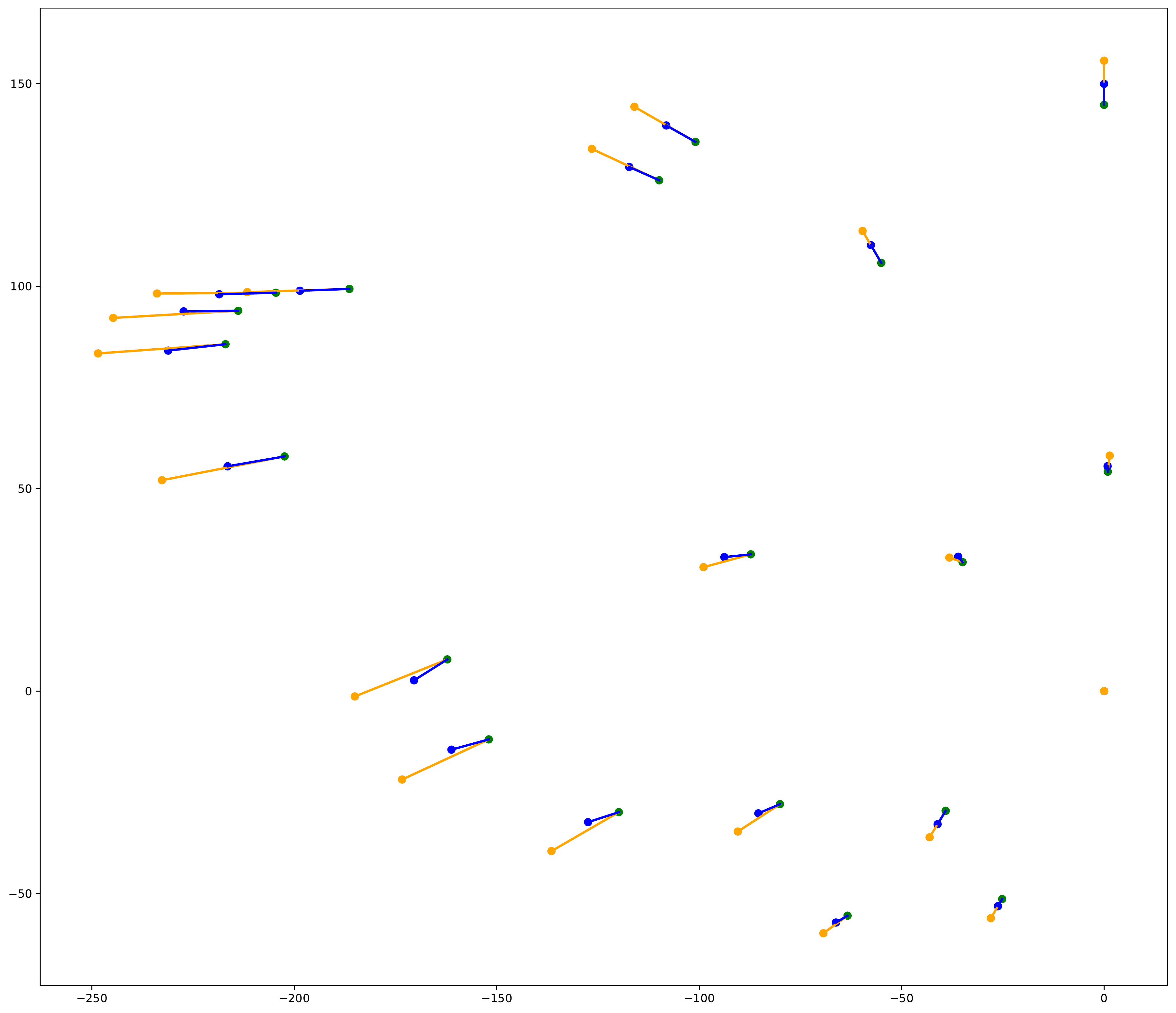}
         \caption{}
         \label{subfig:downwardGrowth}
     \end{subfigure}
     \hfill
     \begin{subfigure}[]{0.38\textwidth}
         \centering
         \includegraphics[width=\textwidth]{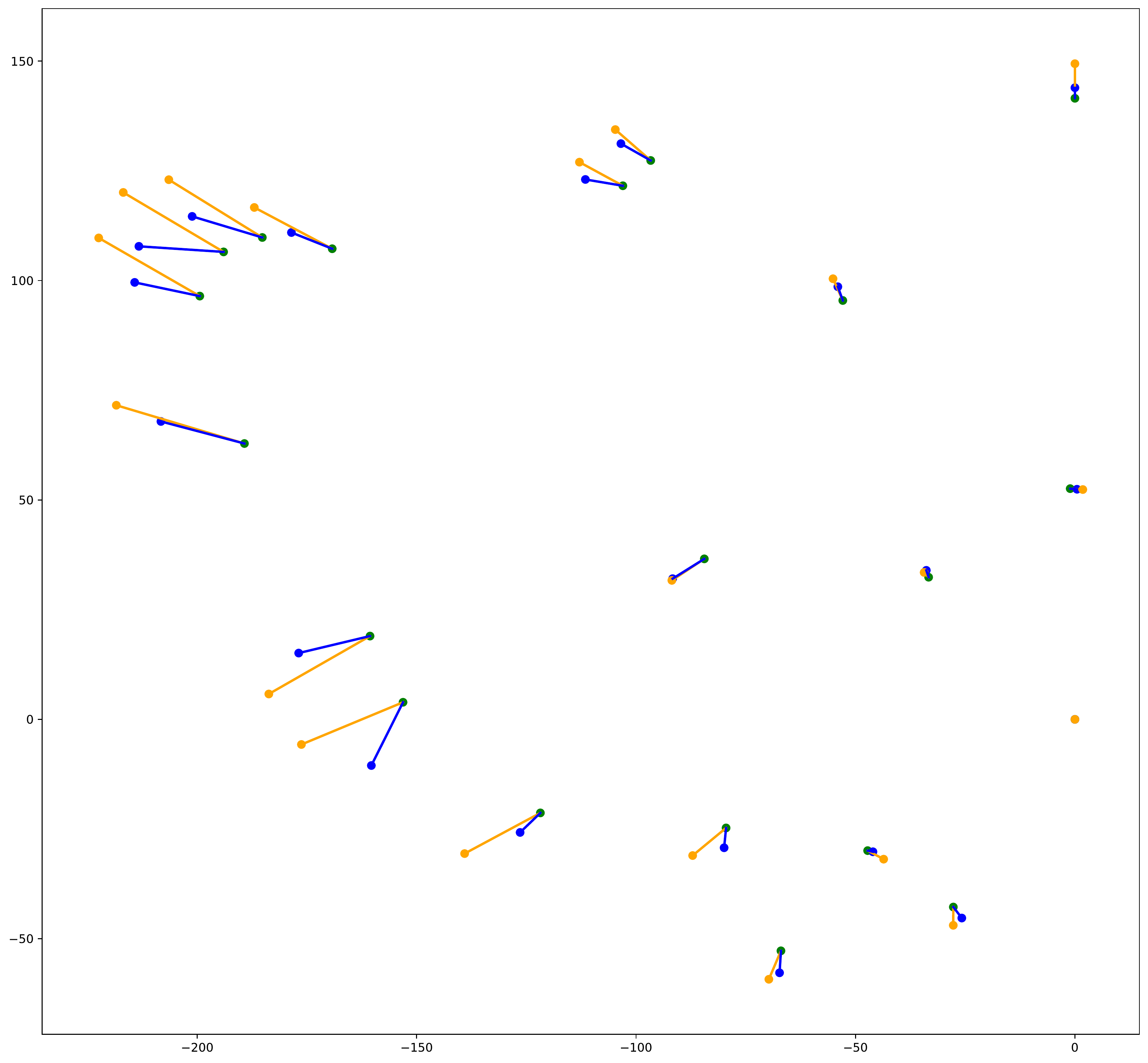}
         \caption{}
         \label{subfig:example1}
     \end{subfigure}
     \begin{subfigure}[]{0.39\textwidth}
         \centering
         \includegraphics[width=\textwidth]{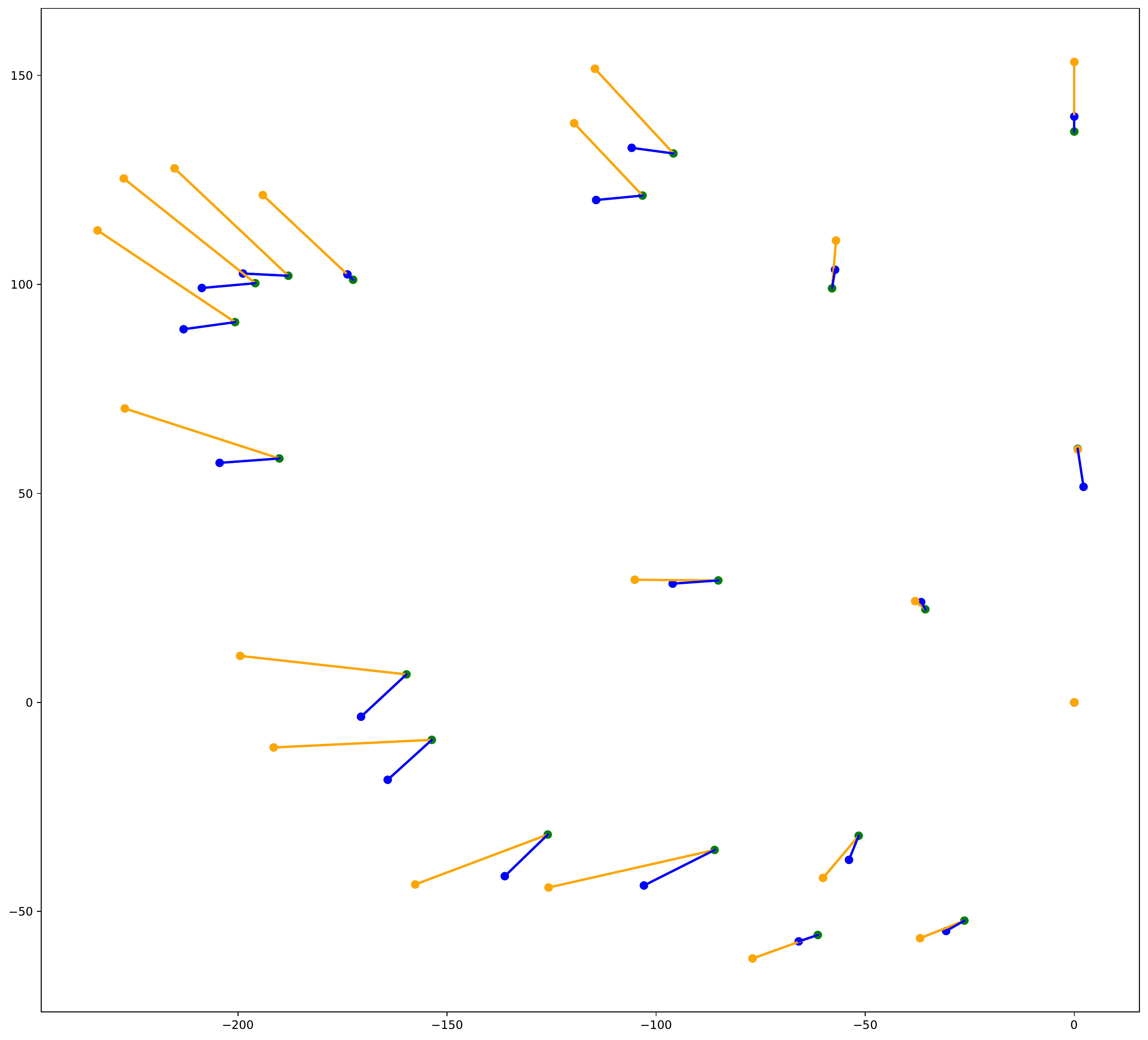}
         \caption{}
         \label{subfig:example2}
     \end{subfigure}
    \caption{Fig. (a) and (b) present average change of landmark coordinates in the group of patients with horizontal (a) and vertical (b) growth. Fig. (c) and (d) illustrate examples of how growth direction between 9--12 may vary from that between 9--18.}
        \label{fig:Growth}
\end{figure}

Next, we determined how patients marked as having horizontal growth differ visually from those labeled as having vertical growth. We selected patients that, according to all three measurements, belonged to the first or third class (for labeling details, please refer to Section~\ref{sec:Experiments}). There were only 31 (out of 639) patients in both categories satisfying all criteria, which shows how ambiguous \emph{horizontal} and \emph{vertical growth} terms are. Fig.~\ref{subfig:forwardGrowth} and Fig.~\ref{subfig:downwardGrowth} present how the average point locations changed along with face growth in both groups. 

Two main conclusions can be drawn. Firstly, the growth directions of all five points belonging to the chin are significantly different for both groups. Secondly, the location of points in the age of 12 is very close to the line set by landmarks in the age of 9 and 18. These two observations looked promising: two classes were well separable and there was high predictive potential thanks to the high 9--12 and 9--18 direction similarity. Naturally, the angle between, for example, the horizontal line and the growth vector between 9 and 18 became a good candidate for a new growth measurement. Unfortunately, when analyzing particular patients, the aforementioned similarity between 9--12 and 9--18 directions is far lower than that of averaged over all patients in a group. Fig.~\ref{subfig:example1} and \ref{subfig:example2} present some examples.

\subsection{Comparison of growth periods}
\label{sec:DataAnalysis:GrowthPeriods}
Having spotted that the growth direction between 9--12 can vary significantly from that between 9--18, we decided to mark additional images of 15-year-olds and measure Pearson correlation of the changes of predicted measurements in particular periods. Table~\ref{tab:correlations} presents the results. Two main conclusions can be drawn. Firstly, there is almost no correlation (Pearson coefficient ranges from -0.37 to 0.05) between three-year periods, which may suggest that growths in particular periods are quite independent of each other. Secondly, the correlation between 9--12 and 9--18 amounts to 0.50, 0.50, and 0.52 for the three predicted measurements.

\begin{table}[b!]
    \centering
    \tiny
    \caption{Correlation of growth in particular periods.}\label{tab:correlations}
        \begin{subtable}{.4\linewidth}
            \caption{SN-MP}
            \begin{tabular}{r|rrrrrr}
            \toprule
            & 9--12 & 12--15 & 15--18 & 9--15 & 12--18 & 9--18 \\
            \hline
            9--12 & 1.00 & & & & & \\					
            12--15 & -0.17 & 1.00 & & & & \\				
            15--18 & -0.10 & 0.01 & 1.00 & & &  \\			
            9--15 &	0.62 & 0.67 & -0.07 & 1.00 & & \\		
            12--18 & -0.20 & 0.81 &	0.59 & 0.50 & 1.00 & \\	
            9--18 &	0.50 & 0.60 & 0.46 & 0.86 &	0.75 & 1.00 \\
            \bottomrule
            \end{tabular}
        \end{subtable}%
        
        \begin{subtable}{.4\linewidth}
            \caption{FA}
            \begin{tabular}{r|rrrrrr}
            \toprule
            & 9--12 & 12--15 & 15--18 & 9--15 & 12--18 & 9--18 \\
            \hline
            9--12 & 1.00 & & & & & \\					
            12--15 & -0.24 & 1.00 & & & & \\				
            15--18 & -0.19 & -0.28 & 1.00 & & &  \\			
            9--15 &	0.63 & 0.60 & -0.38 & 1.00 & & \\		
            12--18 & -0.36 & 0.64 &	0.55 & 0.22 & 1.00 & \\	
            9--18 &	0.50 & 0.40 & 0.35 & 0.73 &	0.63 & 1.00 \\
            \bottomrule
            \end{tabular}
        \end{subtable}%
        
        \begin{subtable}{.4\linewidth}
            \caption{PN-AN}
            \begin{tabular}{r|rrrrrr}
            \toprule
            & 9--12 & 12--15 & 15--18 & 9--15 & 12--18 & 9--18 \\
            \hline
            9--12 & 1.00 & & & & & \\					
            12--15 & -0.18 & 1.00 & & & & \\				
            15--18 & 0.05 & -0.37 & 1.00 & & &  \\			
            9--15 &	0.52 & 0.75 & -0.28 & 1.00 & & \\		
            12--18 & -0.12 & 0.62 &	0.50 & 0.46 & 1.00 & \\	
            9--18 &	0.52 & 0.43 & 0.46 & 0.72 &	0.79 & 1.00 \\
            \bottomrule
            \end{tabular}
        \end{subtable}%
\end{table}

\subsection{Dataset size and class imbalance}
\label{sec:DataAnalysis:DatasetSize}
The number of features ranges from 16 to 82, depending on the experiment, while the number of all samples that have to be split into training, validation, and test sets amounts to 639. Both the low number of instances and the high ratio of features to instances make any deep architecture inapplicable due to overfitting. Thus, we are limited to shallow networks and standard ML methods. Class imbalance, described in more detail in the next section, makes the learning process even more difficult.

%%%%%%%%%%%%%%%%%%%% Experiments %%%%%%%%%%%%%%%%%%%%
\section{Experiments}
\label{sec:Experiments}
Predicted variables are defined for each of the three measurements, SN/MP, FA, and PN-AN, by subtracting the value of a particular measurement at the age of 9 from the corresponding value at the age of 18. Each variable is categorized in the following way:
\begin{itemize}
\item first class is created by instances whose value is lower than one standard deviation from the mean (\emph{horizontal growth});
\item second, most numerous, class contains samples located not further than one standard deviation from the mean (\emph{mixed growth});
\item third class constitutes instances greater than one standard deviation from the mean (\emph{vertical growth}).
\end{itemize}
As a result, for SN-MP, FA, and PN-AN, the majority class contains $68.23\%$, $69.95\%$, and $74.80\%$ instances, respectively. Let us denote the predicted variables as SN-MP(18-9), FA(18-9), and PN-AN(18-9).

\subsection{Models parametrization}
\label{sec:Experiments:ModelsParametrization}
For the sake of brevity of the presentation, let us introduce the following notation:
MLP, MLP($n$), MLP($n_1, n_2$) denote a perceptron with no hidden layer, one hidden layer containing $n$ neurons, and two hidden layers with $n_1$ and $n_2$ neurons, respectively. NN($k$) relates to nearest neighbors classifier considering $k$ neighbors. XGB($r$) refers to the XGBoost algorithm with $r$ boosting rounds. RF($t$) pertains to random forest consisting of $t$ trees. SVM corresponds to Support Vector Machine classifier. LR concerns logistic regression with the maximum number of iterations set to $2\,000$. DT denotes a decision tree.

In all experimental setups defined by the input features (described below) and predicted measurement, the following models were tested:
% MLP, MLP($20$), MLP($50$), MLP($100$), MLP($50, 10$), MLP($50, 20$), MLP($50, 50$), XGB($100$), XGB($300$), RF($100$), RF($300$), SVM, LR, DT, NN($3$), NN($5$).
MLP, MLP(20), MLP(50), MLP(100), MLP(50, 10), MLP(50, 20), MLP(50, 50), XGB(100), XGB(300), RF(100), RF(300), SVM, LR, DT, NN(3), NN(5).

Each hidden layer in neural networks applied ReLU while the last layer -- softmax. The batch size was equal to the number of training instances and we set the maximum number of epochs to 10\,000. 20\% of training samples served as a validation set. In the case of 50 consecutive epochs with no improvement on the validation loss function, training was stopped and weights from the best epoch were restored. The Adam optimizer~\cite{kingma2014adam} was used to optimize the cross-entropy loss function.
The remaining hyperparameters were set to their default values used in Keras 2.3.1 in terms of MLP and in the scikit-learn 0.23.2 in the case of any other model. For each configuration, 5-fold stratified cross-validation was repeated 20 times yielding 100 results altogether.

\subsection{Experimental setup}
\label{sec:Experiments:ExperimentalSetup}
During our experiments, we considered three types of input data: \emph{cephalometric} (ceph), \emph{Procrustes} (proc), and \emph{transformed coordinates} (trans). All of them have a common base which are cephalometric landmarks marked on the 2D X-ray images. Cephalometric data contains mainly angles. Since raw landmark coordinates come from images in different scales, variously rotated and translated, we normalized them so that each meets the following requirements: 
{\bf(1)} a centroid of all landmarks is located at (0, 0); {\bf(2)} a sum of distances between (0, 0) and all transformed coordinates is equal to one; {\bf(3)}
a sum of squared distances between a particular landmark and its average location is minimized across all landmarks and images.
% \begin{itemize}
%     \item a centroid of all landmarks is located at (0, 0);
%     \item a sum of distances between (0, 0) and all transformed coordinates is equal to one;
%     \item a sum of squared distances between a particular landmark and its average location is minimized across all landmarks and images.
% \end{itemize}

\begin{figure}[b!]
\centering
\includegraphics[width=0.78\textwidth]{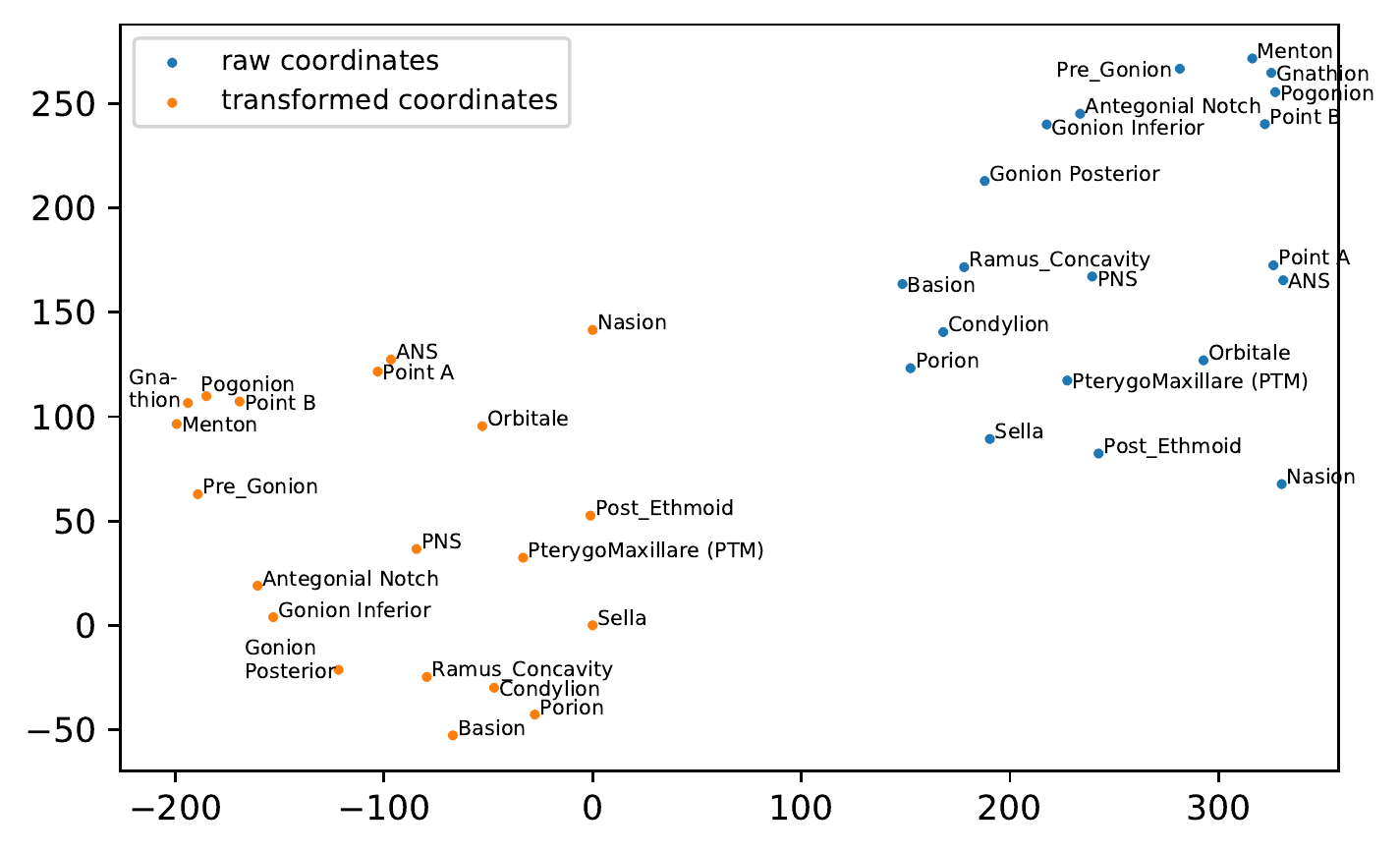}
\caption{Raw and transformed landmarks.}
\label{fig:transformedCoordinates}
\end{figure}

Landmarks formed in the above way are called Procrustes coordinates and are commonly used in biological sciences\cite{zelditch2012geometric}. However, the potential weakness of such representation is the lack of an anchor (all coordinates are relative). Thus, we proposed the third type of feature input which we called transformed coordinates. First, all raw coordinates are moved so that the Sella landmark is located at (0, 0). Then rotation was performed to make Sella-Nasion line vertical. The main reason to unify the Sella-Nasion location is its far distance from the jaw, which presumably allows us to better notice changes during its growth process. Fig.~\ref{fig:transformedCoordinates} depicts raw landmark coordinates and transformed landmarks according to the aforementioned rules along with their names. The number of features related to cephalometric, Procrustes, and transformed data from one timestamp are 15, 40, and 40, respectively. Moreover, the age value is always appended to the feature set.

One of the aims of our experiments was to check whether and how much do we gain by employing data from both the 9th or 12th year of age in comparison to data from only one timestamp (age of 9 or 12). Thus, we created a set of five input feature variants: data at the age of 9 (9), data at the age of 12 (12), difference between the values at the age of 12 and 9 (12-9), data at the ages of 9 and 12 (9, 12), data at the age of 9 and a difference between values at the age of 12 and 9 (9, 12-9).

As mentioned earlier, we predicted three different growth measurements: the change of SN-MP, FA, and PN-AN, each of them between the age of 9 and 18. Tying this all together, 45 experiment scenarios (three types of data, five different periods, and three predicted variables) were formed, for which we run models defined in Section~\ref{sec:Experiments:ModelsParametrization}.

\subsection{Results}
\label{sec:Experiments:Results}
Table~\ref{tab:mainResults} presents five models along with the utilized feature type that obtained the best classification results for each predicted variable and input period.
Exceeding the percentage of the most frequent class (MFC) became a challenge in the case of many configurations. We decided to treat MFC as a baseline (Zero Rule). To examine whether the mean accuracy was statistically higher than the MFC, we ran One Sample \textit{t}-Test and marked with green background results for which \textit{p}-value was less than $0.05$. We denote such results as statistically significant scores (SSS).

\begin{table}[t]
\tiny
\centering
\caption{Best classification results for each predicted variable and input period.}\label{tab:mainResults}

\begin{tabular}{m{0.35cm}m{2.2cm}m{2.2cm}m{2.2cm}m{2.2cm}m{2.2cm}}
\toprule
 & 9 & 12 & 12-9 & 9, 12 & 9, 12-9\\
\midrule
\multirow{2}{0.35cm}{No.} & Features, model & Features, model & Features, model & Features, model & Features, model \\
& Accuracy [\%] & Accuracy [\%] & Accuracy [\%]& Accuracy [\%]& Accuracy [\%] \\
\midrule
\multicolumn{6}{c}{Prediction of SN-MP(18-9)} \\
\midrule
\multirow{2}{0.35cm}{1} & ceph, SVM & ceph, SVM & \cellcolor[HTML]{E2EFDA}ceph, LR & \cellcolor[HTML]{E2EFDA}ceph, LR & \cellcolor[HTML]{E2EFDA}ceph, RF(300)\\
& $68.23 \pm 0.33$ & $68.23 \pm 0.33$ & \cellcolor[HTML]{E2EFDA}$71.25 \pm 2.31$ & \cellcolor[HTML]{E2EFDA}$ 69.12 \pm 2.94 $ & \cellcolor[HTML]{E2EFDA}$ 70.38 \pm 2.25 $ \\ [0.09cm]
\multirow{2}{0.35cm}{2} & proc, SVM & proc, SVM & \cellcolor[HTML]{E2EFDA} ceph, MLP & ceph, SVM & \cellcolor[HTML]{E2EFDA}ceph, RF(100)\\
& $68.23 \pm 0.33$ & $68.23 \pm 0.33$ & \cellcolor[HTML]{E2EFDA} $70.07 \pm 2.84$ & $68.23 \pm 0.33$ & \cellcolor[HTML]{E2EFDA}$ 70.19 \pm 2.45 $ \\ [0.09cm]
\multirow{2}{0.35cm}{3} & trans, SVM & trans, SVM &\cellcolor[HTML]{E2EFDA} ceph, RF(300) & proc, SVM & \cellcolor[HTML]{E2EFDA}ceph, XGB(100) \\
& $68.23 \pm 0.33$ & $68.23 \pm 0.33$ & \cellcolor[HTML]{E2EFDA} $70.06 \pm 2.21$ & $68.23 \pm 0.33$ &\cellcolor[HTML]{E2EFDA} $ 69.90 \pm 2.84 $ \\ [0.09cm]
\multirow{2}{0.35cm}{4} & proc, MLP & proc, LR & \cellcolor[HTML]{E2EFDA}ceph, RF(100) & trans, SVM & \cellcolor[HTML]{E2EFDA}ceph, XGB(300) \\
& $68.22 \pm 0.35$ & $68.23 \pm 0.33$ & \cellcolor[HTML]{E2EFDA}$69.77 \pm 2.33$ & $68.23 \pm 0.33$ & \cellcolor[HTML]{E2EFDA}$ 69.59 \pm 2.87 $ \\ [0.09cm]
\multirow{2}{0.35cm}{5} & proc, MLP(50, 50) & proc, MLP & \cellcolor[HTML]{E2EFDA}proc, XGB(100) & proc, RF(300) & \cellcolor[HTML]{E2EFDA}ceph, LR \\ 
& $68.21 \pm 0.39$ & $68.23 \pm 0.41$ & \cellcolor[HTML]{E2EFDA}$69.06 \pm 2.50$ & $ 68.14 \pm 0.79$ & \cellcolor[HTML]{E2EFDA}$69.06 \pm 2.92 $ \\
\midrule
\multicolumn{6}{c}{Prediction of FA(18-9)} \\
\midrule
\multirow{2}{0.35cm}{1} & proc, MLP(50, 50) & proc, MLP & \cellcolor[HTML]{E2EFDA}proc, MLP(50, 20) & proc, MLP(20) & \cellcolor[HTML]{E2EFDA}proc, MLP(50, 10)\\
& $ 69.98 \pm 0.42 $ & $69.97 \pm 0.37$ & \cellcolor[HTML]{E2EFDA}$71.12 \pm 2.41 $ & $70.05 \pm 1.18$ & \cellcolor[HTML]{E2EFDA}$70.92\pm 2.26 $ \\ [0.09cm]
\multirow{2}{0.35cm}{2} & ceph, SVM & ceph, SVM & \cellcolor[HTML]{E2EFDA}proc, MLP(50, 50) & ceph, SVM & \cellcolor[HTML]{E2EFDA}proc, MLP(50, 20) \\
& $ 69.95 \pm 0.35 $ & $69.95 \pm 0.35$ & \cellcolor[HTML]{E2EFDA}$71.01 \pm 2.45$ & $69.95 \pm 0.35$ & \cellcolor[HTML]{E2EFDA}$70.74 \pm 2.28 $ \\ [0.09cm]
\multirow{2}{0.35cm}{3} & proc, SVM & proc, SVM & \cellcolor[HTML]{E2EFDA}proc, MLP(20) & proc, SVM & \cellcolor[HTML]{E2EFDA}proc, MLP(50) \\
& $ 69.95 \pm 0.35 $ & $69.95 \pm 0.35$ & \cellcolor[HTML]{E2EFDA}$70.99 \pm 2.09$ & $69.95 \pm 0.35$ & \cellcolor[HTML]{E2EFDA}$70.69 \pm 2.24$ \\ [0.09cm]
\multirow{2}{0.35cm}{4} & trans, SVM & trans, SVM & \cellcolor[HTML]{E2EFDA}proc, MLP(50) & trans, SVM & \cellcolor[HTML]{E2EFDA}proc, MLP(20) \\
& $ 69.95 \pm 0.35 $ & $69.95 \pm 0.35$ & \cellcolor[HTML]{E2EFDA}$70.99 \pm 2.35$ & $69.95 \pm 0.35$ & \cellcolor[HTML]{E2EFDA}$70.66 \pm 1.88$ \\ [0.09cm]
\multirow{2}{0.35cm}{5} & proc, LR & proc, LR & \cellcolor[HTML]{E2EFDA}ceph, LR & proc, LR & \cellcolor[HTML]{E2EFDA}proc, MLP(100)\\
& $ 69.95 \pm 0.35 $ & $69.95 \pm 0.35$ & \cellcolor[HTML]{E2EFDA}$70.95 \pm 2.86 $ & $69.95 \pm 0.35$ & \cellcolor[HTML]{E2EFDA}$70.64 \pm 2.11$ \\
\midrule
\multicolumn{6}{c}{Prediction of PN-AN(18-9)} \\
\midrule
\multirow{2}{0.35cm}{1} & \cellcolor[HTML]{E2EFDA}ceph, LR & \cellcolor[HTML]{E2EFDA}ceph, LR & ceph, LR & trans, RF(300) & ceph, SVM\\
& \cellcolor[HTML]{E2EFDA}$75.05 \pm 0.94$ & \cellcolor[HTML]{E2EFDA}$75.25 \pm 1.14$ & $75.14 \pm 2.06$ & $74.99 \pm 1.58$ & $74.80 \pm 0.30$ \\ [0.09cm]
\multirow{2}{0.35cm}{2} & trans, RF(300) & trans, RF(300) & ceph, MLP & trans, RF(100) & proc, SVM\\
& $74.95 \pm 1.46$ & $74.95 \pm 1.50$ & $74.97 \pm 2.23$ & $74.88 \pm 1.64$ & $74.80 \pm 0.30$ \\ [0.09cm]
\multirow{2}{0.35cm}{3} & trans, RF(100) & ceph, SVM & proc, SVM & ceph, LR & proc, LR \\
& $74.91 \pm 1.49$ & $74.80 \pm 0.30$ & $74.80 \pm 0.30$ & $74.82 \pm 2.05$ & $74.80 \pm 0.30$ \\ [0.09cm]
\multirow{2}{0.35cm}{4} & ceph, SVM & proc, SVM & proc, LR & ceph, SVM & trans, RF(300)\\
& $74.80 \pm 0.30$ & $74.80 \pm 0.30$ & $74.80 \pm 0.30$ & $74.80 \pm 0.30$ & $74.80 \pm 1.70$ \\ [0.09cm]
\multirow{2}{0.35cm}{5} & proc, SVM & proc, LR & trans, SVM & proc, LR & ceph, LR \\
& $74.80 \pm 0.30$ & $74.80 \pm 0.30$ & $74.75 \pm 0.37$ & $74.80 \pm 0.30$ & $74.78 \pm 2.17$ \\
\bottomrule
\end{tabular}
\end{table}

The first observation is that among all configurations with input from one timestamp, only two models achieved SSS and it was just $0.25$ pp and $0.45$ pp above MFC, respectively. It leads us to the vital conclusion that data coming from both timestamps, the ages of 9 and 12 (or their difference), are essential to predict face growth. 

Second, the difference between the corresponding values at the age of 12 and 9 constitutes the best feature set. For both predicted variables, SN-MP(18-9) and FA(18-9), all five models built on such features achieved SSS. Since those models obtained better accuracy than models with additional features being values at the age of 9 or models built on values from the age of 9 and 12, we inferred that crucial for the final prediction is a change between 9 and 12 rather than values at particular timestamps themselves.   

Third, all models that achieved SSS were built on cephalometric or Procrustes features. We presume that the main reason behind the lack of a model based on transformed coordinates is the non-uniform image scale. 

Next, PN-AN(18-9) is the hardest to predict. Only two algorithms achieved SSS, $75.25 \pm 1.14\%$ and $75.05 \pm 0.94\%$, for that variable. There is one more model with a bit higher mean accuracy than the latter, 75.14\%, but due to substantial standard deviation it is considered as not statistically significant.

The final observation is that all SSS were obtained by logistic regression, some types of multilayer perceptron or tree ensemble -- random forest or XGBoost.

MFC was slightly exceeded by our models for all three predicted measurements. It could seem to be a trivial achievement. However, one should be aware that the more problem is imbalanced, the harder is to beat MFC. For the sake of the next experiment, we artificially relabeled our data so that each class constitutes one-third of the total number of samples (instances with the lowest and highest value of the predicted variable from the majority class were tagged as a first or third class). Best models achieved an accuracy ranging from $51\%$ to $55\%$, depending on the predicted variable. It showed that in the case of evenly distributed classes, our models significantly exceeded MFC, which amounted to $33\%$ in this setting.

%%%%%%%%%%%%%%%%%%%% Conclusions and future work %%%%%%%%%%%%%%%%%%%%
\section{Conclusions and future work}
\label{sec:Conclusions}
To our knowledge, this is the first paper that handles the problem of FG prediction with ML methods. The task turned out to be quite challenging. Among the root factors that make the problem so difficult, we pointed different sources of data, different tools taking X-ray photographs, diversified image scale, which is not possible to determine, ambiguous landmarking, low number of instances, and class imbalance. Finally, we were limited to 2D cephalograms since we could not utilize 3D radiographic images due to ethical issues. 

Data analysis showed that on a statistical level, the growth directions in periods 9--12, 12--15, and 15--18 are very lowly correlated. Nevertheless, our models achieved statistically significant higher accuracy than the percentage of the most frequent class. Overcoming the aforesaid issues may presumably increase model performance. However, given the number and scope of conducted experiments, we hypothesise that 
%having such kind of data, it is impossible to increase the results significantly. It would mean that 
the information essential to predict FG lies beyond the landmark coordinates of 9- and 12-year-olds.

We identified two main lines of inquiry worth pursuing in the future. The first one pertains to further work on tabular data. In this scope, we plan to apply some data augmentation methods to enlarge the size and variety of the dataset along with some feature selection that serves as a denoiser. Moreover, we want to analyze some other face normalization methods. Finally, together with orthodontic experts, we plan to think of new FG measurements that would be easier to predict. 

The second line of further research relates to processing raw images. Certainly, due to the nuanced character of the problem, we do not expect it to be an easy task as well. Nevertheless, we hope that some models, especially CNNs (presumably shallow due to a low number of training instances), which are widely used in many computer vision tasks, will be able to find some patterns that are imperceptible for the human eye.

\paragraph{Acknowledgments} Studies were funded by BIOTECHMED-1 project granted by Warsaw University of Technology under the program Excellence Initiative: Research University (ID-UB). 

\bibliographystyle{splncs04}
\bibliography{references}

\begin{thebibliography}{10}
\providecommand{\url}[1]{\texttt{#1}}
\providecommand{\urlprefix}{URL }
\providecommand{\doi}[1]{https://doi.org/#1}

\bibitem{american1996growth}
{AGS} (1996), \url{https://aaoflegacycollection.org/}, online; accessed
  02-Jun-2021

\bibitem{arik2017fully}
Arik, S.{\"O}., Ibragimov, B., Xing, L.: {Fully automated quantitative
  cephalometry using convolutional neural networks}. Journal of Medical Imaging
   \textbf{4}(1),  014501 (2017)

\bibitem{baumrind1984prediction}
Baumrind, S., Korn, E.L., West, E.E.: {Prediction of mandibular rotation: an
  empirical test of clinician performance}. AJO  \textbf{86}(5),  371--385
  (1984)

\bibitem{bhatia1979proposed}
Bhatia, S., Wright, G., Leighton, B.: {A proposed multivariate model for
  prediction of facial growth}. AJO  \textbf{75}(3),  264--281 (1979)

\bibitem{bjork1969prediction}
Bj{\"o}rk, A.: {Prediction of mandibular growth rotation}. AJO  \textbf{55}(6),
   585--599 (1969)

\bibitem{buschang1990mandibular}
Buschang, P., Tanguay, R., LaPalme, L., Demirjian, A.: {Mandibular growth
  prediction: mean growth increments versus mathematical models}. EJO
  \textbf{12}(3),  290--296 (1990)

\bibitem{el1994longitudinal}
El-Batouti, A., {\O}gaard, B., Bishara, S.E.: Longitudinal cephalometric
  standards for norwegians between the ages of 6 and 18 years. EJO
  \textbf{16}(6),  501--509 (1994)

\bibitem{hwang2020automated}
Hwang, H.W., Park, J.H., Moon, J.H., Yu, Y., Kim, H., Her, S.B., Srinivasan,
  G., Aljanabi, M.N.A., Donatelli, R.E., Lee, S.J.: {Automated identification
  of cephalometric landmarks: Part 2-Might it be better than human?} The Angle
  Orthodontist  \textbf{90}(1),  69--76 (2020)

\bibitem{kingma2014adam}
Kingma, D.P., Ba, J.: {Adam: A Method for Stochastic Optimization}. arXiv
  preprint arXiv:1412.6980  (2014)

\bibitem{kolodziej2002evaluation}
Kolodziej, R.P., Southard, T.E., Southard, K.A., Casko, J.S., Jakobsen, J.R.:
  {Evaluation of antegonial notch depth for growth prediction}. AJO-DO
  \textbf{121}(4),  357--363 (2002)

\bibitem{lux1999evaluation}
Lux, C.J., Conradt, C., Stellzig, A., Komposch, G.: {Evaluation of the
  predictive impact of cephalometric variables. Logistic regression and ROC
  curves}. Journal of Orofacial Orthopedics/Fortschritte der
  Kieferorthop{\"a}die  \textbf{60}(2),  95--107 (1999)

\bibitem{park2019automated}
Park, J.H., Hwang, H.W., Moon, J.H., Yu, Y., Kim, H., Her, S.B., Srinivasan,
  G., Aljanabi, M.N.A., Donatelli, R.E., Lee, S.J.: {Automated identification
  of cephalometric landmarks: Part 1—Comparisons between the latest
  deep-learning methods YOLOV3 and SSD}. The Angle Orthodontist
  \textbf{89}(6),  903--909 (2019)

\bibitem{perillo2000effect}
Perillo, M., Beideman, R., Shofer, F., Jacobsson-Hunt, U., Higgins-Barber, K.,
  Laster, L., Ghafari, J.: Effect of landmark identification on cephalometric
  measurements: guidelines for cephalometric analyses. Clin Ortho Res
  \textbf{3}(1),  29--36 (2000)

\bibitem{rudolph1998multivariate}
Rudolph, D.J., White, S.E., Sinclair, P.M.: {Multivariate prediction of
  skeletal Class II growth}. AJO-DO  \textbf{114}(3),  283--291 (1998)

\bibitem{skieller1984prediction}
Skieller, V., Bj{\"o}rk, A., Linde-Hansen, T.: {Prediction of mandibular growth
  rotation evaluated from a longitudinal implant sample}. AJO  \textbf{86}(5),
  359--370 (1984)

\bibitem{zelditch2012geometric}
Zelditch, M.L., Swiderski, D.L., Sheets, H.D.: Geometric Morphometrics for
  Biologists: A Primer. Academic Press (2004)

\end{thebibliography}

\end{document}